%% file: main.tex
\pdfoutput=1

\documentclass[11pt]{article}

\usepackage[]{EMNLP2023}

\usepackage{times}
\usepackage{latexsym}

\usepackage{amsmath}
\usepackage{amsthm}
\usepackage{booktabs}
\usepackage{multirow}
\usepackage{graphicx}
\usepackage{threeparttable}

\usepackage{enumitem}
\usepackage{float}
\usepackage{amsfonts}
\usepackage[ruled,vlined]{algorithm2e}
\usepackage{algorithmic}
\usepackage{url}

\usepackage[T1]{fontenc}

\usepackage[utf8]{inputenc}

\usepackage{microtype}

\usepackage{inconsolata}

\usepackage{graphicx}

\title{
Enhancing Short-Text Topic Modeling with LLM-Driven Context Expansion and Prefix-Tuned VAEs
}

\author{Pritom Saha Akash $\quad$ Kevin Chen-Chuan Chang \\
University of Illinois at Urbana-Champaign, USA \\
 \texttt{\{pakash2, kcchang\}@illinois.edu}
}

\begin{document}
\maketitle

\input{sections/abstract}
\input{sections/introduction}

\input{sections/related}
\input{sections/method}

\input{sections/experiments}

\input{sections/conclusion}

\input{sections/acknowledgements}
\input{sections/limitation}
\bibliography{anthology}
\bibliographystyle{acl_natbib}

\input{sections/appendix}

\end{document}

%% file: sections/abstract.tex
\begin{abstract}
Topic modeling is a powerful technique for uncovering hidden themes within a collection of documents. However, the effectiveness of traditional topic models often relies on sufficient word co-occurrence, which is lacking in short texts. Therefore, existing approaches, whether probabilistic or neural, frequently struggle to extract meaningful patterns from such data, resulting in incoherent topics. To address this challenge, we propose a novel approach that leverages large language models (LLMs) to extend short texts into more detailed sequences before applying topic modeling. To further improve the efficiency and solve the problem of semantic inconsistency from LLM-generated texts, we propose to use prefix tuning to train a smaller language model coupled with a variational autoencoder for short-text topic modeling. Our method significantly improves short-text topic modeling performance, as demonstrated by extensive experiments on real-world datasets with extreme data sparsity, outperforming current state-of-the-art topic models. \footnote{Code and data are available at \href{https://github.com/pritomsaha/PVTM}{https://github.com/ pritomsaha/PVTM}}

\end{abstract}

%% file: sections/introduction.tex
\section{Introduction}
\label{sec:introduction}

In the digital era, short texts like tweets, web page titles, news headlines, image captions, and product reviews are prevalent for sharing knowledge. However, the sheer volume of these texts necessitates efficient information extraction mechanisms. Topic modeling is a key method for uncovering latent topics in short texts, with applications including comment summarization \cite{ma2012topic}, content characterization \cite{ramage2010characterizing,zhao2011comparing}, emergent topic detection \cite{lin2010pet}, document classification \cite{sriram2010short}, user interest profiling \cite{weng2010twitterrank}, and so on.

Traditional topic models, such as LDA and PLSA, are designed to uncover latent topics given a corpus of documents by analyzing word co-occurrences within the texts \cite{blei2003latent, hofmann1999probabilistic}. These models assume that each document contains enough text to provide meaningful co-occurrence information. However, in short texts like titles, captions, and headlines, this assumption does not hold due to the limited text in each document. This scarcity leads to a data sparsity problem, where limited word co-occurrences make it difficult for traditional models to effectively mine high-quality topics. The primary challenge is that each document is short, rather than the corpus being insufficient in size.

While various strategies have been developed for modeling topics in short texts, each has its limitations. E.g., aggregating short texts into longer pseudo-documents based on metadata like user information, hashtags, or external corpora is a common approach \citet{weng2010twitterrank,mehrotra2013improving, zuo2016topic}; however, the availability of such metadata can be inconsistent. To overcome this, some methods rely on structural or semantic information within the texts themselves, such as the Biterm Topic Model \cite{yan2013biterm} and its extensions \cite{zhu2018graphbtm}, which focus on word pairs but often cannot provide individual document topic distributions. Another method \citet{yin2014dirichlet} limits texts to a single topic, simplifying the model but potentially overlooking texts that span multiple topics.

\input{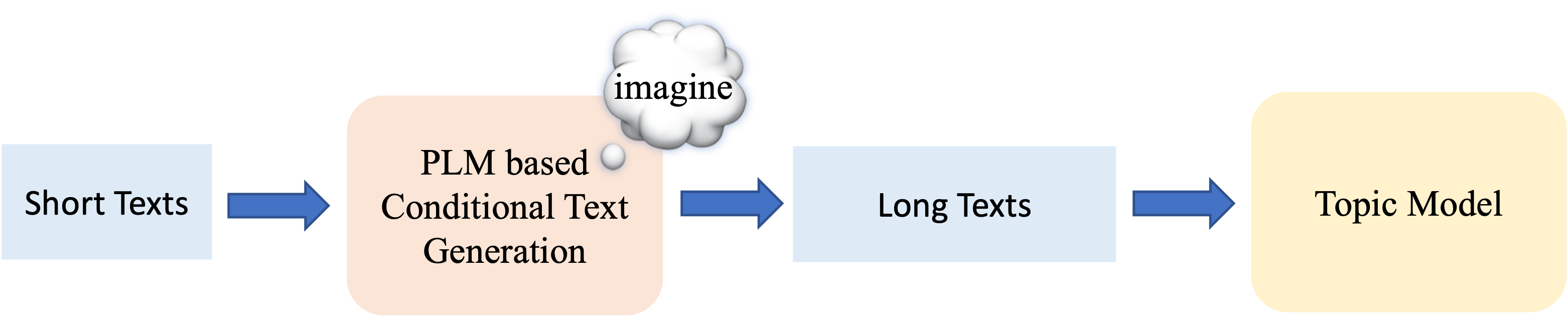}

Considering the limitations mentioned above, in this paper, we first try to understand the characteristics of short texts and how humans process these texts when detecting topics. A short text, such as a title or caption, typically serves as a summarized version of a longer text, providing readers with essential hints about the full content. When judging the topics of short texts, humans often infer the broader context based on their background knowledge and the cues provided in the text. For example, given the headline: "No tsunami but FIFA's corruption storm rages on," readers might use their understanding of "FIFA" to infer that the headline pertains to the topic of "sports."

This leads us to the question: Can a model similarly infer the broader context to better understand the topics of a short text? Recently, large language models (LLMs) such as GPT-3 \cite{brown2020language}, LLAMA2 \cite{touvron2023llama}, and T5 \cite{raffel2020exploring, chung2022scaling} have demonstrated remarkable capabilities as open-ended text generators, capable of producing surprisingly fluent text from a limited preceding context. For example, given the abovementioned news headline, LLMs can generate extended sequences (as shown in the third and fourth columns of Table \ref{tab:error_example} with tokens such as "FIFA World Cup" and "Soccer," which are strongly related to the sport of soccer. This ability to generate contextually relevant information suggests that LLMs can be leveraged to enrich the contextual information of short texts, thereby improving topic modeling.

Considering these capabilities, we first explore the potential solution for short-text topic modeling: leveraging large language models (LLMs) to generate a longer text from each short text in a corpus before applying traditional topic modeling techniques. By expanding short texts into more detailed, context-rich narratives, LLMs can create a proxy for the detailed context that traditional topic modeling techniques often lack when dealing with short texts. In other words, it is a proxy of human-like inference of the broader context surrounding a given short text before mining the topics, as shown in Figure \ref{fig:arch0}.


While leveraging LLMs to expand short texts offers a promising solution, this approach faces several \textbf{challenges}. One key challenge is maintaining \textit{semantic consistency}: ensuring that the generated longer texts accurately reflect the original short texts without introducing irrelevant or inaccurate information is difficult, as LLMs are not always fine-tuned for specific tasks or domains. This can lead to semantic drift, distorting the topic modeling results. Another challenge is the \textit{latency} of LLM calls: generating extended texts online can be slow, making real-time topic detection impractical, even if offline generation during training is feasible.

To tackle these challenges, we aim to avoid directly using LLM-generated longer texts as input. Instead, we train a model to learn topics from short texts and reconstruct longer texts previously generated by an LLM. This minimizes the effects of any shift in meaning in the generated texts. By decoding topics from short texts before generating longer texts, we align with one of the LLM's inherent characteristics. As noted by \cite{wang2023large}, LLMs implicitly engage in topic modeling by navigating a latent conceptual space to generate text, with each token generation influenced by an underlying topic variable. However, directly inferring these latent concepts into discrete topics like Latent Dirichlet Allocation (LDA) is not straightforward.

To bridge this gap, we introduce the \textit{Prefix-tuned Variational Topic Model} (PVTM), which combines a smaller language model (LM) with a variational autoencoder (VAE) for topic inference. Instead of tuning the entire LM, we employ prefix tuning \cite{li2021prefix}, which fine-tunes only a small set of parameters, effectively capturing domain-specific features from short texts. This minimizes the effects of any semantic drift in the generated texts from larger, general LLMs. The extracted features from smaller LM serve as input for a VAE to decode discrete topics. The features extracted from the smaller LM are then used by the VAE to decode discrete topics. Both the LM and VAE are trained together with a topic modeling objective to ensure effective learning.

The key insights of our solution include -- (1) Semantic Consistency: By training on short texts and using generated longer texts only as output, we ensure the integrity of the original data and mitigate the risk of introducing irrelevant information. (2) Efficiency: The reduced inference time of smaller LMs and the efficiency of VAEs in learning discrete topics make this method suitable for real-time topic detection applications. (3) Prefix Tuning: This fine-tuning method allows us to capture domain-specific features without the computational overhead of tuning large LLMs, ensuring scalability.

To summarize, our \textbf{contributions} in this paper are the following. Firstly, we explore LLMs for extending short texts into longer ones and then apply traditional topic models to the longer texts. Secondly, to improve efficiency and address the issue of semantic drift, we propose a new framework consisting of a jointly trained smaller LM and VAE. Finally, we conduct a comprehensive set of experiments on multiple datasets across different tasks, demonstrating our model's superiority over existing baselines.
\input{tables/examples}

%% file: figures/architecture0.tex
\begin{figure*}[!tb]
\centering
\includegraphics[width=0.9\linewidth]{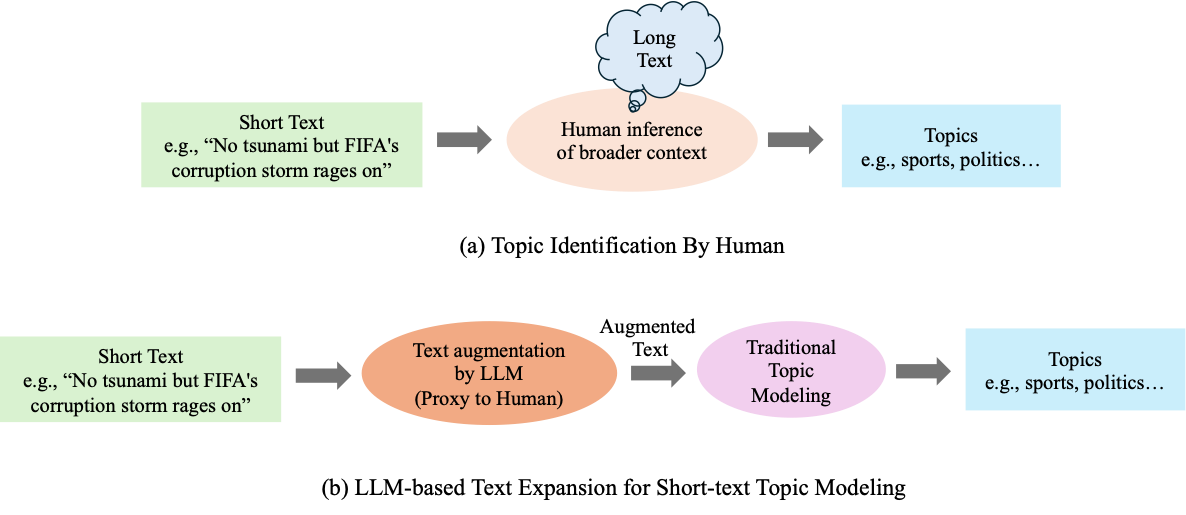}
\caption{LLM for Short-Text Topic Modeling}
\label{fig:arch0}
\vspace{-3mm}
\end{figure*}

%% file: tables/examples.tex
\begin{table*}[]
\resizebox{1.0\linewidth}{!}{%
\begin{tabular}{@{}lc@{}}
\toprule
\textbf{Prompt}                                                           & \begin{tabular}[c]{@{}l@{}}Given the short text: "\{Short Text\}", expand it into a detailed paragraph that provides background and elaborates \\ on the key points to enrich its context. Try to make it as detailed as possible.\end{tabular}                                                                                                                                                                                                                                                                                                                                                                                                                                                                                                                                                                                                                                                                                                                    \\ \midrule
\textbf{Short Text}                                                       & No tsunami but FIFA's corruption storm rages on                                                                                                                                                                                                                                                                                                                                                                                                                                                                                                                                                                                                                                                                                                                                                                                                                                                                                                                    \\ \midrule
\begin{tabular}[c]{@{}l@{}}\textbf{Extended Text}\\ (T5-XXL)\end{tabular} & \begin{tabular}[c]{@{}l@{}}After years of corruption scandals, FIFA is in the grip of an even greater one. A year after FIFA's biggest corruption \\ scandal in its history, the world soccer governing body continues to roil with revelations, and the WTA has called for \\ the re-establishment of a special committee to investigate the scandal. A year after the FIFA Ethics Committee launched \\ its investigation into allegations of corruption in the sport, the investigation is still going strong, with new allegations \\ surfacing on a daily basis. After the FIFA World Cup corruption scandal, a wave of resignations has swept across the \\ organization. A tsunami of scandals engulfed FIFA as it awaited a decision on the replacement of the new president. \\ A year after FIFA's most serious corruption scandal broke, the corruption storm continues to rage with FIFA chiefs.\end{tabular}                                          \\ \midrule
\begin{tabular}[c]{@{}l@{}}\textbf{Extended Text}\\ (LLAMA2)\end{tabular} & \begin{tabular}[c]{@{}l@{}}Despite the absence of a tsunami, the corruption storm surrounding FIFA, the governing body of international soccer, \\ continues to rage on. The organization has been embroiled in a series of scandals over the past few years, with allegations \\ of bribery, money laundering, and other forms of corruption among top officials and sponsors. The most recent controversy \\ involves the awarding of the 2022 World Cup hosting rights to Qatar, which has been marred by allegations of vote-buying \\ and other unethical practices. The fallout from these scandals has been significant, with several top FIFA officials arrested and \\ charged with criminal offenses, and the organization's reputation taking a hit as a result. Despite efforts to clean up its act, the \\ corruption storm continues to rage on, with many calling for greater transparency and accountability within the organization.\end{tabular} \\ \bottomrule
\end{tabular}
}
\caption{Example short text and corresponding extended texts using two different LLMs.}
\label{tab:error_example}
\vspace{-5mm}
\end{table*}

%% file: sections/related.tex
\section{Related Work}
\label{sec:related_work}

\subsection{Traditional Topic Models}
Traditional probabilistic topic models like Probabilistic Latent Semantic Analysis (PLSA) \cite{hofmann1999probabilistic} and Latent Dirichlet Allocation (LDA) \cite{blei2003latent} work well with large-sized documents, relying on ample co-occurrence information to capture latent topic structures. However, these models often struggle with short texts such as news titles and image captions. To address this, the Biterm Topic Model (BTM) \cite{yan2013biterm} utilizes structural and semantic information, while another strategy aggregates short texts into longer pseudo-documents using metadata (e.g., hashtags, external corpora) before applying conventional topic models \cite{mehrotra2013improving,zuo2016topic}. Another approach, the Dirichlet Multinomial Mixture (DMM) model \cite{yin2014dirichlet,nigam2000text}, assumes each document is sampled from a single topic. Although intuitive, this assumption can be overly restrictive as many short texts may cover multiple topics.

\subsection{Neural Topic Models}
With the recent developments in deep neural networks (DNNs) and deep generative models, there has been an active research direction in leveraging DNNs for inferring topics from corpus, also called neural topic modeling. The recent success of variational autoencoders (VAE) \cite{kingma2013auto} has opened a new research direction for neural topic modeling \cite{nan2019topic}. The first work that uses VAE for topic modeling is called the Neural Variational Document Model (NVDM) \cite{miao2016neural}, which leverages the reparameterization trick of Gaussian distributions and achieves a fantastic performance boost. Another related work called ProdLDA \cite{srivastava2017autoencoding} uses Logistic Normal distribution to handle the difficulty of the reparameterization trick for Dirichlet distribution. 

There also have been several works in neural topic modeling (NTM) for short texts. E.g., \cite{zeng2018topic} combines NTM with a memory network for short text classification. \cite{zhu2018graphbtm} takes the idea of the probabilistic biterm topic model to NTM where the encoder is a graph neural network (GNN) of sampled biterms. However, this model is not generally able to generate the topic distribution of an individual document. \cite{lin2020copula} introduce the Archimedean copulas idea in the neural topic model to regularise the discreteness of topic distributions for short texts, which restricts the document from some salient topics. From a similar intuition, \cite{feng2022context} proposes an NTM by limiting the number of active topics for each short document and also incorporating the word distributions of the topics from pre-trained word embeddings. Another neural topic model \cite{wu2020short} employs a topic distribution quantization approach to generate peakier distributions that are better suited to modeling short texts.

\subsection{LMs in Topic Models}
Previous neural topic models have used language models (LMs) to represent documents. For example, the contextualized topic model (CTM) \cite{bianchi2020pre} combines a document's Bag of Words (BOW) representation with its contextualized vector from LMs like BERT \cite{devlin2018bert}, capturing context and order information that BOW misses. Similarly, BERTopic \cite{grootendorst2022bertopic} uses LM-based document embeddings for clustering and TF-IDF to identify representative words as topics. However, BERTopic's reliance on TF-IDF doesn't fully utilize LMs' ability to capture word semantics. DeTime \cite{xu-etal-2023-detime} improves clusterability and semantic coherence by using Encoder-Decoder-based LLMs for embeddings. Despite these advances, these models don't address the data sparsity issue in short-text topic modeling; they only improve document representation for general-purpose topic modeling. In contrast, our proposed framework leverages LMs for conditional text generation to enrich the contextual information of short documents.

%% file: sections/method.tex
\section{Proposed Methodology}
\label{sec:method}
Our proposed framework consists of two components. The first component generates longer text given a short text. The second one utilizes the generated longer texts for topic modeling.

\subsection{Short Text Extension}
\label{sec:short_exten}
As specified before, according to \cite{wang2023large}, LLMs inherently perform topic modeling. This is achieved by treating each token generation as a decision informed by a latent topic or concept variable $\theta$, suggesting that LLMs understand and generate text by navigating a latent conceptual space. More specifically, LLMs generate new tokens based on all previous tokens $ P(w_{1:T}) = \prod_{i=1}^{T} P(w_i|w_{i-1}, \ldots, w_1) $ and it  can be decomposed as below:
\begin{align*}
   &P_M(w_{t+1:T}|w_{1:t}) \\
   &= \int_{\Theta} P_M(w_{t+1:T}|\theta)P_M(\theta|w_{1:t})d\theta 
\end{align*}
where $M$ is a specific LLM.
This illustrates the LLM's process of generating text conditioned on previous tokens and a latent topic variable, integrating over all possible conceptual themes $\Theta$ that could inform the generation. However, we can not explicitly obtain the latent concept variable to understand the topic. Therefore, we formulate the short text extension as a conventional conditional sentence generation task, i.e., generating longer text sequences given a short text. Formally, we use the standard sequence-to-sequence generation formulation with an LLM $\mathcal{M}$: given input a short text sequence $x$, the probability of the generated long  sequence $y = [y_1, \dots, y_m]$ is calculated as:
\begin{align*}
    \mathbf{Pr}_{\mathcal{M}}(y|x) = \sum_{i=1}^m \mathbf{Pr}_{\mathcal{M}}(y_i|y_{<i}, x),
\end{align*}
where $y_{<i}$ denotes the previous tokens $y_1,\dots, y_{i-1}$. The LLM $\mathcal{M}$ specific text generation function $f_{\mathcal{M}}$ is used for sampling tokens and the sequence with the largest $ \mathbf{Pr}_{\mathcal{M}}(y|x)$ probability is chosen. Later, we use the extended text to decode the inherent topic in LLMs.

\subsection{Topic Model on Generated Long Text} 
\label{sec:LCSNTM}
Upon obtaining the longer text sequences from the previous step, one straightforward approach is to use existing topic models that perform better with long text documents. As the longer texts have better co-occurrence context than the original short texts, it is expected to reduce the data sparsity problem of short-text topic modeling. Thus, exploring existing probabilistic and neural topic models on the generated longer text sequences is intuitive. Therefore, we directly utilize different existing topic models on generated texts as one solution, as shown in Figure \ref{fig:arch0}.

However, directly using LLMs generated text for topic modeling may pose a risk. The generated text might shift from the original domain or only partially cover the intended topics. For example, consider a short text about ``renewable energy sources'':
\begin{itemize}[nolistsep,leftmargin=*]
    \item \textit{Original short text:} ``Renewable energy sources like solar and wind power are essential for reducing carbon emissions and combating climate change.''
    \item \textit{ChatGPT-generated longer text \cite{openai2023chatgpt}:}  ``Renewable energy sources, such as solar power and wind turbines, are becoming increasingly popular worldwide. These sources harness natural elements to generate electricity, contributing to the reduction of greenhouse gases. Solar panels capture sunlight and convert it into energy, while wind turbines use the wind's kinetic energy. Additionally, hydroelectric power, geothermal energy, and biomass are also crucial renewable sources. Countries are investing heavily in these technologies to transition from fossil fuels to cleaner energy solutions.''
\end{itemize}
While the generated text provides a detailed overview of various renewable energy sources, it introduces new topics like hydroelectric power, geothermal energy, and biomass. This expansion can be beneficial for providing a broader context but may deviate from the original focus on solar and wind power. The opposite scenario is also possible, where the original short text is about multiple topics, and the generated long text is missing some of these topics, leading to incomplete topic coverage in a document.



To solve this issue, we propose a solution called Prefix-tuned Variational Topic Model (PVTM), as shown in Figure~\ref{fig:arch1}.

\input{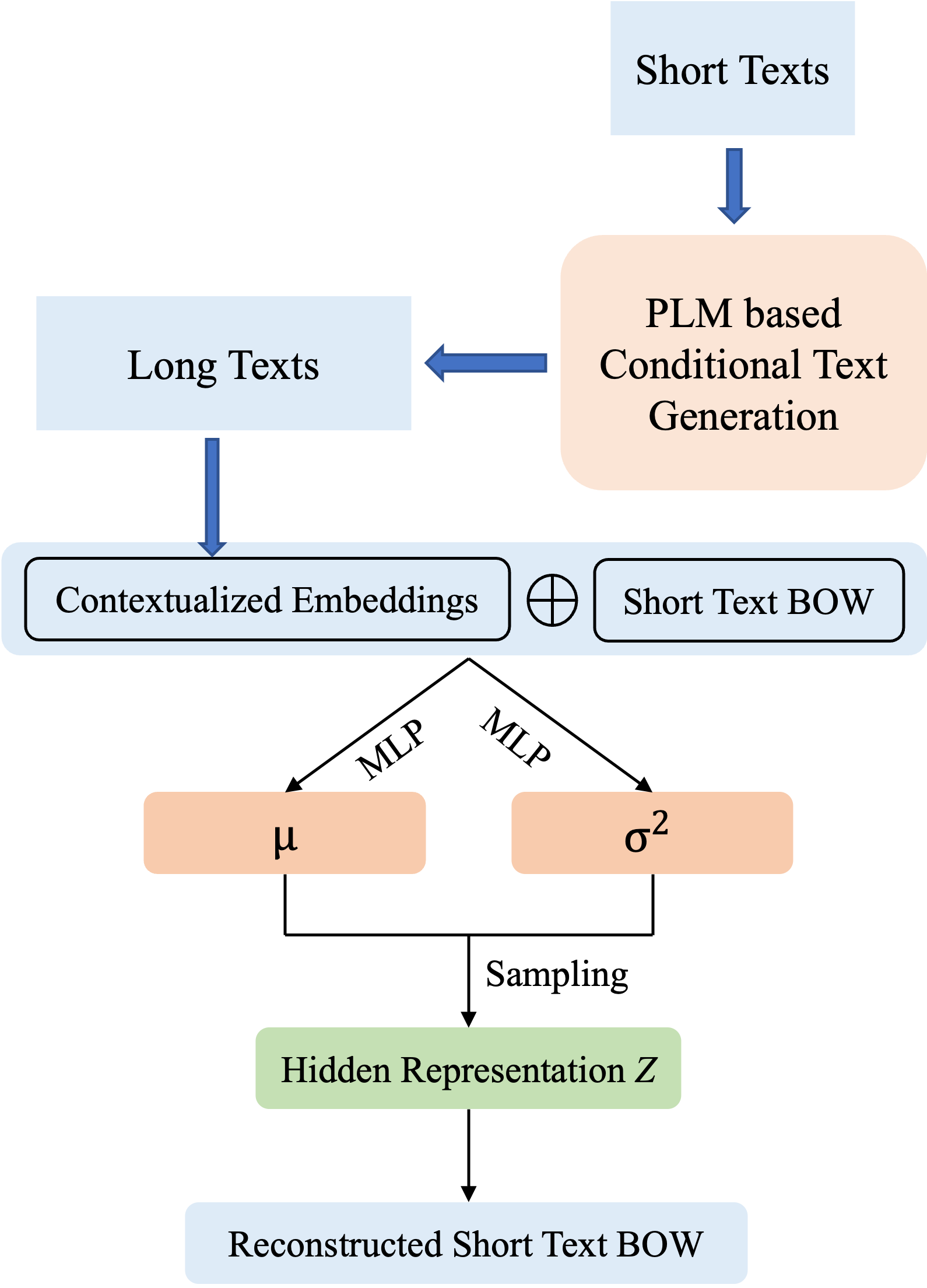}
{\flushleft \textbf{PVTM:}}
To address the issues of deviations from the original focus or incomplete topic coverage in LLM-generated long texts as input, we propose using the generated sequences solely as outputs, reconstructed from the short texts rather than inputs to the topic model. Formally, our model builds on the existing neural topic model ProdLDA \cite{srivastava2017autoencoding}, which is based on the Variational AutoEncoder (VAE) mechanism \cite{kingma2013auto}. ProdLDA uses a variational inference network to map the bag-of-words (BOW) representation of a document to a continuous latent space. However, BOW representations fail to capture important semantic nuances, especially in short texts where the context is limited. To overcome this, we replace the BOW input with a smaller language model (LM) to encode short texts, thereby learning richer, task-specific features relevant to topic modeling. 

Training an entire LM for this task can be computationally expensive, particularly when dealing with large-scale corpora or real-time applications. Furthermore, fully fine-tuning all parameters of a large LM is often unnecessary, as many pre-trained weights already encapsulate general linguistic knowledge. Therefore, to improve efficiency while still adapting the model to our specific task, we use Prefix Tuning \cite{li2021prefix}, a parameter-efficient fine-tuning approach. The core idea is to prepend trainable vectors (prefixes) to the input embeddings at each transformer layer, allowing the LM to adjust its behavior for the task without modifying its main pre-trained weights. 

Let $\mathbf{X}$ be the original input embeddings of a short text, and $\mathbf{P}$ be the trainable prefix vectors. The combined input to each transformer layer is then $\mathbf{X}^{\prime} = [\mathbf{P}, \mathbf{X}]$, where $\mathbf{X}^{\prime}$ represents the concatenation of the prefix $\mathbf{P}$ with the input $\mathbf{X}$. These prefix vectors are optimized along with the task objective, allowing the model to adapt to specific tasks like topic modeling with minimal parameter updates. Formally, for each transformer layer, the attention mechanism operates on $\mathbf{X}^{\prime}$:
$
\text{Attention}(\mathbf{Q}, \mathbf{K}, \mathbf{V}) = \text{softmax}\left(\frac{\mathbf{Q}\mathbf{K}^T}{\sqrt{d}}\right)\mathbf{V}
$, where the query $\mathbf{Q}$, key $\mathbf{K}$, and value $\mathbf{V}$ matrices now incorporate the prefix information, allowing the transformer layers to attend to task-specific signals provided by the prefixes. The parameters of these prefixes are updated during training while the rest of the language model remains frozen, ensuring that task-specific knowledge is encoded without retraining the entire model.

The output of the LM, enhanced by prefix tuning, is then fed into the VAE for topic inference. Specifically, the model first generates a mean vector $\mu$ and a variance vector $\sigma^2$ through two separate MLPs from a document. The $\mu$ and $\sigma^2$ are then used to sample a latent representation $Z$ assuming a Gaussian distribution. Subsequently, a decoder network reconstructs the BOW representation of the extended long texts generated by LLMs by generating words from $Z$. The model is trained with the original objective function \cite{srivastava2017autoencoding} called the evidence lower bound (ELBO), defined as follows:

\resizebox{0.9\linewidth}{!}{
\begin{minipage}{\linewidth}
\vspace{-4mm}
\begin{align}
    \mathcal{L}(\Theta) = \sum_{d\in \mathcal{D}} \sum_{n=1}^{N_d} \mathbb{E}_q[\log p(w_{dn} \mid Z_d)] - \nonumber \\
    \sum_{d\in \mathcal{D}} KL (q(Z_d;w_d, \Theta) \mid \mid p(Z_d)),
    \label{eq:elbo}
\end{align}
  \end{minipage}
}
where $w_{dn}$ is the $n$-th token in a document $d$ with length $N_d$ from the corpus $\mathcal{D}$. $\Theta$ represents learnable parameters in the model. $q(\cdot)$ is a Gaussian whose mean and variance are estimated from two separate MLPs.

%% file: figures/architecture1.tex
\begin{figure}[!tb]
\centering
\includegraphics[width=0.6\linewidth]{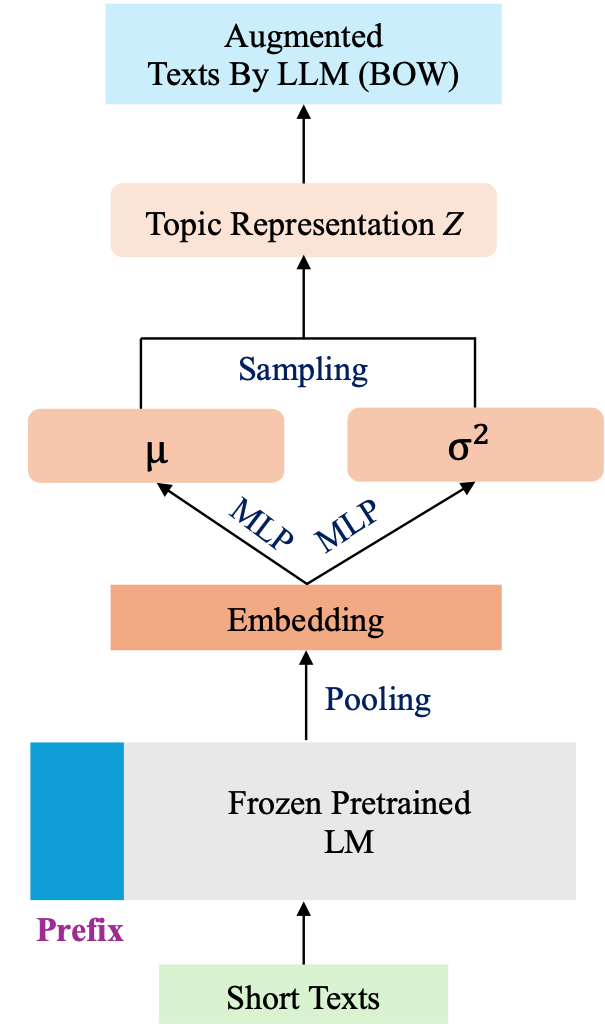}
\caption{Proposed Architecture of PVTM}
\label{fig:arch1}
\vspace{-5mm}
\end{figure}

%% file: sections/experiments.tex
\section{Experiments}
\label{sec:experiment}
In this section, we employ empirical evaluations, which are designed mainly to fulfill the following objectives:

\begin{itemize}[nolistsep,leftmargin=*]

\item Does the LLMs grounded text extension improve the performance of existing topic models?
 
\item How effectively does the proposed PVTM improve the performance of topic modeling for short texts?

\item How qualitatively different are the topics discovered by the proposed architecture from existing baselines?

\end{itemize} 

\subsection{Experiment Setup}

{\flushleft \textbf{Datasets.}} We use the following datasets to evaluate our proposed architecture. The detailed statistics of these datasets are shown in Table \ref{tab:dataset_stat}.
\begin{itemize}[nolistsep,leftmargin=*]

    \item \textbf{TagMyNews:} Titles and contents of English news articles published by \citet{vitale2012classification} are included in this dataset . In our experiment, we use the headlines from the news as brief paragraphs. Every news item is given a ground-truth name, such as ``sci-tech'', ``business'', etc.

    \item \textbf{Google News:} The web content from Google search snippets makes up the dataset provided by \citet{yin2014dirichlet}. It is a snapshot of Google News on November 27, 2013.  It includes the titles and brief descriptions of 11,108 news articles, which are organized into 152 distinct categories or clusters.
    
    \item \textbf{StackOverflow:} 
     This dataset was created using the challenge information that was provided in Kaggle\footnote{\url{https://www.kaggle.com/datasets/stackoverflow/stackoverflow}}. 
     We make use of the dataset which contains 20,000 randomly chosen question titles. Information technology terms like ``matlab'', ``osx'', and ``visual studio'' are labeled next to each question title.

\end{itemize}

\input{tables/dataset_stat}

{\flushleft \textbf{Baselines.}}
We compare our models with the following baselines.
\begin{itemize}[nolistsep,leftmargin=*]
    \item \textbf{LDA:} We used one of the widely used probabilistic topic models, Latent Dirichlet Allocation (LDA) \cite{blei2003latent} as a baseline for this work.

    \item \textbf{ProdLDA:} This is the very first paper that adapted variation auto-encoders for topic modeling \cite{srivastava2017autoencoding}.

    \item \textbf{NMF:} Negative Matrix Factorization \cite{wang2012nonnegative} based topic modeling.
    \item \textbf{BERTopic:} BERTopic \cite{grootendorst2022bertopic} uses transformers and c-TF-IDF to create interpretable topics by forming dense clusters that preserve important words in descriptions.


\item \textbf{NQTM:} A state-of-the-art neural short text topic
model with vector quantization. \cite{wu2020short}

\item \textbf{CTM:} Contextualized Topic Model combines contextualized representations of documents with neural topic models \cite{bianchi2020pre}. 
    
\item \textbf{CLNTM}: Contrastive Learning for Neural Topic Model combines contrastive learning paradigm with neural topic models by considering both effects of positive and negative pairs \cite{nguyen2021contrastive}.

\item \textbf{TSCTM}: It is another constrastive learning-based approach that uses quantization for better positive and negative sampling. \cite{wu2022mitigating}.

\item \textbf{vONTSS}: This method \cite{xu-etal-2023-vontss}  introduces a semi-supervised neural topic modeling method that leverages von Mises-Fisher (vMF) variational autoencoders and optimal transport to optimize topic-keyword quality and classification using a small set of keywords per topic.

\item \textbf{DeTime}: DeTime \cite{xu-etal-2023-detime} leverages encoder-decoder-based large language models (LLMs) to produce highly clusterable embeddings that generate topics with superior clusterability and enhanced semantic coherence.

\end{itemize}
For extending short texts into longer ones, we use LLAMA2 \cite{touvron2023llama}. 

The implementation details are discussed in Appendix~\ref{sec:imp_details}.

\input{tables/topic_quality}

\subsection{Topic Quality Evaluation} 

{\flushleft \textbf{Evaluation Metrics.}}
For evaluating the quality of topics returned by each model, we use the following two different metrics:
\begin{itemize}[nolistsep,leftmargin=*]

 \item \textbf{$C_V$}: We use the widely used coherence score for topic modeling named $C_V$. It is a standard measure of the interpretability of topics \cite{wu2020short}.

 \item \textbf{$IRBO$}: Inverted Rank-Biased Overlap (IRBO) evaluates the topic diversity by calculating rank-biased overlap over the generated topics \cite{webber2010similarity}. 
\end{itemize}

{\flushleft \textbf{Results and Discussions.}}
We first analyze the result of existing topic models on the generated text from an LLM (described in Section \ref{sec:method}). The topic quality scores ($C_V$, and IRBO) in Table~\ref{tab:topic_quality_new} show the apparent dominance of topic models on extended text compared to short texts. The best NPMI and IRBO scores for all three datasets are from extended texts with significant improvement in topic coherency and comparable diversity. This clearly shows that the extension of short text using LLMs helps discover higher-quality topics that are more coherent and diverse. For example, in LDA, while using extended texts, the coherence score $C_V$ improves from 0.399 to 0.523 compared to short texts.

However, these topic quality results do not always show that the mined topics correctly represent the target dataset. As specified in Section \ref{sec:LCSNTM}, the topics may shift because of the LLM-generated texts. We further discuss this through classification results in the next section. Now, considering the topic quality performance of the proposed PVTM, we identify some interesting findings. In almost all cases, we get an improvement in topic quality scores compared to both the short-texts and extendeded texts counterparts . More specifically, we obtained a significant performance boost in terms of coherence and diversity scores compared to all other baselines. E.g., in the TagMyNews dataset, compared to the most similar model CTM,  the $C_V$ score for PVTM increases from 0.618 to 0.632 (for K=20 topics). 

In Appendix \ref{sec:ablation}, we also report and discuss the topic quality results of different variants of PVTM based on the types of inputs and outputs.

\subsection{Text Classification Evaluation}
Although text classification is not the main purpose of topic models, the generated document topic distribution can be used as the document feature for learning text classifiers. Therefore, we evaluate how learned document topic distribution is distinctive and informative enough to represent a document to be used for classifying a document correctly. We employ two different classification models on top of document topic distribution learned by different models. The classification models are Support Vector Machine (SVM) \cite{cortes1995support} and Logistic Regression (LR) \cite{wright1995logistic}. We use classification accuracy over 5-fold cross-validation to compare the performance of multiple classifiers.

\input{tables/classification}
\input{tables/qualitative_result}

{\flushleft \textbf{Results and Discussions.}}
The classification result is presented in Table \ref{tab:classification}. Overall, the proposed PVTM is the best-performing model regarding classification accuracy, leveraging both the generated text and considering the topics shift (or incomplete coverage of topics) problem. As specified before, when using LLMs without finetuning on the target corpus, the generated text may not cover the original topics of the document or shift from them. Even if the StackOverflow dataset is about a particular technical domain, the LLMs are more likely to generate tokens from general domains. That is why the learned topics from the extended texts may not represent the original documents, resulting in poor classification performance.
This effect is comparatively less in the other two datasets, as those are about more general topics like ``politics'', ``sports'', etc. On the other hand, the PVTM reduces this effect by using the original short texts as input during training, which is also visible in the classification result.  

\vspace{-2mm}
\subsection{Topic Examples Evaluation}

To evaluate the proposed models qualitatively, we show the top 10 words for each of the three topics generated by different models in Table \ref{tab:qualitative}. We observe that some models on short texts generate topics with repetitive words (e.g., CLNTM). Although the CTM on short texts generates diverse topics, they are less informative and coherent (i.e., confluencing multiple topics like iOS and general applications, etc.). On the other hand, topics in generated long texts are less repetitive with much more coherency, although some also tend to generate topics with general words like ``number'' and ``size''. Finally, the PVTM generates both non-repetitive and informative topics. E.g., it is easy to detect that the three discovered topics are database, shell, and web programming.

%% file: tables/dataset_stat.tex
\begin{table}[]
\resizebox{1.0\linewidth}{!}{%
\begin{tabular}{@{}ccccc@{}}
\toprule
Datasets      & \# of docs & \begin{tabular}[c]{@{}c@{}}Average\\ length\end{tabular} & \begin{tabular}[c]{@{}c@{}}\# of class\\ labels\end{tabular} & \begin{tabular}[c]{@{}c@{}}Vocabulary\\ size\end{tabular} \\ \midrule
TagMyNews Titles     & 5000       & 5.78                                                     & 7                                                            & 7111                                                      \\
Google News   & 11108       & 6.11                                                    & 152                                                            & 7187                                                     \\ 
StackOverflow & 19899      & 4.49                                                     & 20                                                           & 8556                                                      \\
\bottomrule
\end{tabular}}
\caption{Statistics of datasets after preprocessing.}
\label{tab:dataset_stat}
\vspace{-1mm}
\end{table}

%% file: tables/topic_quality.tex
\begin{table*}[!ht]
\centering
\resizebox{0.9\linewidth}{!}{%
\begin{tabular}{@{}ll|cccccccccccc@{}}
\hline
\multicolumn{2}{l}{\multirow{3}{*}{Method}} & \multicolumn{4}{c}
{TagMyNews Titles}                                                                            & \multicolumn{4}{c}{Google News}                                                                                 & \multicolumn{4}{c}{StackOverflow}                                                                               \\ \cline{3-14} 
\multicolumn{2}{c}{}                        & \multicolumn{2}{c}{K=20}                               & \multicolumn{2}{c}{K=50}                               & \multicolumn{2}{c}{K=20}                               & \multicolumn{2}{c}{K=50}                               & \multicolumn{2}{c}{K=20}                               & \multicolumn{2}{c}{K=50}                               \\
\multicolumn{2}{c}{}                        & \multicolumn{1}{c}{$C_V$} & \multicolumn{1}{c}{$IRBO$} & \multicolumn{1}{c}{$C_V$} & \multicolumn{1}{c}{$IRBO$} & \multicolumn{1}{c}{$C_V$} & \multicolumn{1}{c}{$IRBO$} & \multicolumn{1}{c}{$C_V$} & \multicolumn{1}{c}{$IRBO$} & \multicolumn{1}{c}{$C_V$} & \multicolumn{1}{c}{$IRBO$} & \multicolumn{1}{c}{$C_V$} & \multicolumn{1}{c}{$IRBO$} \\ \midrule
\multirow{2}{*}{LDA}            &\textit{ST}       & 0.399                     & 0.981                      & 0.369                     & 0.983                      & 0.326                     & 0.996                      & 0.347                     & 0.998                      & 0.413                     & 0.980                      & 0.396                     & 0.991                      \\
                                &\textit{ET}       & 0.523                     & 0.979                      & 0.498                     & 0.989                      & 0.414                     & 0.990                       & 0.433                     & 0.991                      & 0.501                     & 0.638                      & \textbf{0.492}            & 0.935                      \\ \midrule

\multirow{2}{*}{ProdLDA}           &\textit{ST}       & 0.439                     & 0.984                      & 0.410                     & 0.991                      & 0.417                     & 0.997                      & 0.391                     & 0.996                      & 0.495                     & 0.977                      & 0.446                     & 0.977                      \\

                                &\textit{ET}       
                                & 0.587 & 0.985 & 0.583 & 0.990 & 0.498 & 0.989 & 0.488 & 0.993 & 0.474 & 0.978 & 0.473 & 0.980 
                                \\ \midrule

\multirow{2}{*}{NMF}           &\textit{ST}      & 0.439 & 0.984 & 0.410 & 0.991 & 0.417 & 0.997 & 0.391 & 0.996 & 0.495 & 0.969 & 0.446 & 0.977 \\

                                &\textit{ET}       
                                & 0.587 & 0.985 & 0.583 & 0.990 & 0.498 & 0.989 & 0.488 & 0.993 & 0.474 & 0.978 & 0.473 & 0.980 
                                \\ \midrule

\multirow{2}{*}{BERTopic}           &\textit{ST}      & 0.584 & 0.996 & 0.516 & 0.998 & 0.345 & 0.998 & 0.380 & 0.998 & 0.446 & 0.980 & 0.392 & 0.984 \\

                                &\textit{ET}       
                                & 0.614 & 0.976 & 0.559 & 0.989 & 0.423 & 0.995 & 0.429 & 0.996 & 0.500 & 0.930 & 0.457 & 0.973 
                                \\ \midrule
\multirow{2}{*}{NQTM}           &\textit{ST}       & 0.322                     & 0.941                      & 0.345                     & 0.937                      & 0.258                     & 0.973                      & 0.289                     & 0.942                      & 0.291                     & 0.993                      & 0.327                     & 0.991                      \\
                                &\textit{ET}       & 0.542                     & 1.000                          & 0.551                     & 0.999                      & 0.405                     & 1.000                          & 0.438                     & 1.000                          & 0.301                     & 1.000                          & 0.218                     & 1.000                          \\ \midrule
\multirow{2}{*}{CTM}            &\textit{ST}       & 0.481                     & 1.000                      & 0.531                     & 0.991                      & 0.351                     & 1.000                      & 0.393                     & 0.994                      & 0.410                     & 1.000                      & 0.392                     & 0.986                      \\
                                &\textit{ET}       & 0.618                     & 0.997                      & 0.566                     & 0.991                      & 0.421                     & 0.988                      & 0.472                     & 0.995                      & 0.411                     & 0.994                      & 0.437                     & 0.990                       \\ \midrule
\multirow{2}{*}{CLNTM}          &\textit{ST}       & 0.311                     & 0.972                      & 0.356                     & 0.942                      & 0.324                     & 0.995                      & 0.356                     & 0.942                      & 0.324                     & 0.995                      & 0.296                     & 0.845                      \\
                                &\textit{ET}       & 0.613                     & 0.988                      & 0.541                     & 0.979                      & \textbf{0.503}            & 0.999                      & 0.513                     & 0.994                      & 0.412                     & 0.998                      & 0.438                     & 0.990                       \\ \midrule
\multirow{2}{*}{TSCTM}          &\textit{ST}       & 0.363                     & 1.000                      & 0.304                     & 1.000                      & 0.284                     & 1.000                      & 0.298                     & 1.000                      & 0.124                     & 1.000                      & 0.121                     & 0.997                      \\
                                &\textit{ET}       & 0.585                     & 1.000                          & 0.391                     & 1.000                          & 0.35                      & 1.000                          & 0.338                     & 1.000                          & 0.151                     & 1.000                          & 0.108                     & 1.000                          \\ \midrule
\multirow{2}{*}{vONTSS}           &\textit{ST}       & 0.409                     & 0.788                      & 0.397                     & 0.930                       & 0.349                     & 0.981                      & 0.348                     & 0.933                      & 0.281                     & 0.723                      & 0.358                     & 0.868                      \\
                                &\textit{ET}       & 0.536                     & 0.994                      & 0.457                     & 0.983                      & 0.418                     & 0.999                      & 0.404                     & 0.991                      & 0.413                     & 0.998                      & 0.392                     & 0.982                      \\ \midrule
\multirow{2}{*}{DeTime}         &\textit{ST}       & 0.398                     & 0.779                      & 0.403                     & 0.922                      & 0.288                     & 0.719                      & 0.326                     & 0.903                      & 0.279                     & 0.664                      & 0.361                     & 0.849                      \\
                                &\textit{ET}       & 0.427                     & 0.976                      & 0.37                      & 0.963                      & 0.371                     & 0.954                      & 0.32                      & 0.938                      & 0.381                    & 0.797                      & 0.36                      & 0.907                      \\ \midrule
PVTM                           &           & \textbf{0.632}            & \textbf{1.000}             & \textbf{0.585}            & \textbf{1.000}                 & 0.445                     & \textbf{1.000}                 & \textbf{0.452}            & \textbf{1.000}                 & \textbf{0.558}            & \textbf{1.000}                 & 0.462                     & \textbf{1.000}
\\ 
\bottomrule 
\end{tabular}}
\caption{Topic coherences ($C_V$) and diversity ($IRBO$) scores of topic words. $K$ denotes the number of topics. The best result in each case is shown in \textbf{bold}. \textit{ST:} Short Texts, \textit{ET:} Extended Texts}
\label{tab:topic_quality_new}
\end{table*}

%% file: tables/classification.tex
\begin{table*}[!ht]
\centering
\resizebox{0.9\linewidth}{!}{%

\begin{tabular}{@{}ll|cccccccccccc@{}}
\hline
\multicolumn{2}{c}{\multirow{3}{*}{Method}} & \multicolumn{4}{c}
{TagMyNews Titles}                                                                            & \multicolumn{4}{c}{Google News}                                                                                 & \multicolumn{4}{c}{StackOverflow}                                                                               \\ \cline{3-14} 
\multicolumn{2}{l}{}                        & \multicolumn{2}{c}{K=20}                               & \multicolumn{2}{c}{K=50}                               & \multicolumn{2}{c}{K=20}                               & \multicolumn{2}{c}{K=50}                               & \multicolumn{2}{c}{K=20}                               & \multicolumn{2}{c}{K=50}                               \\
\multicolumn{2}{l}{}                        & \multicolumn{1}{c}{SVM} & \multicolumn{1}{l}{LR} & \multicolumn{1}{c}{SVM} & \multicolumn{1}{c}{LR} & \multicolumn{1}{c}{SVM} & \multicolumn{1}{c}{$LR$} & \multicolumn{1}{c}{SVM} & \multicolumn{1}{c}{$LR$} & \multicolumn{1}{c}{SVM} & \multicolumn{1}{c}{$LR$} & \multicolumn{1}{c}{SVM} & \multicolumn{1}{c}{$LR$} \\ \midrule
\multirow{2}{*}{LDA}            &\textit{ST}       & 0.247          & 0.317          & 0.259          & 0.303          & 0.235          & 0.354          & 0.432          & 0.535          & 0.381          & 0.431          & 0.561          & 0.605          \\
                                &\textit{ET}       & 0.695          & 0.718          & 0.725          & 0.737          & 0.292          & 0.531          & 0.529          & 0.737          & 0.522          & 0.588          & 0.658          & 0.707          \\ \midrule

\multirow{2}{*}{ProdLDA}           &\textit{ST}      & 0.410 & 0.438 & 0.396 & 0.432 & 0.361 & 0.629 & 0.587 & 0.805 & 0.614 & 0.675 & 0.768 & 0.777           \\
                                &\textit{ET}      & 0.718 & 0.728 & 0.738 & 0.761 & 0.356 & 0.526 & 0.590 & 0.779 & 0.498 & 0.580 & 0.672 & 0.718          \\ \midrule

\multirow{2}{*}{NMF}           &\textit{ST}      & 0.386 & 0.449 & 0.423 & 0.501 & 0.290 & 0.432 & 0.538 & 0.690 & \textbf{0.745} & 0.769 & 0.803 & 0.766           \\
                                &\textit{ET}      & 0.714 & 0.736 & 0.710 & 0.751 & 0.300 & 0.536 & 0.527 & 0.719 & 0.434 & 0.564 & 0.708 & 0.776          \\ \midrule

\multirow{2}{*}{BERTopic}           &\textit{ST}      & 0.504 & 0.547 & 0.614 & 0.641 & 0.298 & 0.379 & 0.534 & 0.605 & 0.722 & 0.734 & 0.732 & 0.743           \\
                                &\textit{ET}      & 0.718 & 0.729 & 0.737 & 0.747 & 0.308 & 0.469 & 0.525 & 0.676 & 0.717 & 0.725 & 0.803 & 0.801          \\ \midrule

\multirow{2}{*}{NQTM}           &\textit{ST}       & 0.123          & 0.254          & 0.123          & 0.254          & 0.023          & 0.038          & 0.114          & 0.309          & 0.050           & 0.050           & 0.050           & 0.050           \\
                                &\textit{ET}       & 0.172          & 0.249          & 0.188          & 0.241          & 0.013          & 0.037          & 0.011          & 0.028          & 0.049          & 0.054          & 0.048          & 0.055          \\ \midrule
\multirow{2}{*}{CTM}            &\textit{ST}       & 0.595          & 0.619          & 0.668          & 0.694          & 0.283          & 0.512          & 0.514          & 0.679          & 0.705 & 0.739          & 0.814          & \textbf{0.817} \\
                                &\textit{ET}       & 0.686          & 0.721          & 0.736          & 0.757          & 0.339          & 0.547          & 0.592          & 0.762          & 0.462          & 0.58           & 0.656          & 0.719          \\ \midrule
\multirow{2}{*}{CLNTM}          &\textit{ST}       & 0.165          & 0.260           & 0.165          & 0.251          & 0.020           & 0.066          & 0.050           & 0.095          & 0.065          & 0.121          & 0.050           & 0.100            \\
                                &\textit{ET}       & 0.703          & 0.718          & 0.720           & 0.736          & 0.343          & 0.619          & 0.565          & 0.782          & 0.522          & 0.659          & 0.624          & 0.67           \\ \midrule
\multirow{2}{*}{TSCTM}          &\textit{ST}       & 0.423          & 0.473          & 0.485          & 0.527          & 0.337          & 0.518          & 0.498          & 0.685          & 0.565          & 0.736          & 0.774          & 0.784          \\
                                &\textit{ET}       & 0.721          & 0.732          & \textbf{0.755}          & 0.713          & 0.314          & \textbf{0.699} & 0.594          & 0.630           & 0.557          & 0.657          & 0.687          & 0.726          \\ \midrule
\multirow{2}{*}{vONTSS}           &\textit{ST}       & 0.316          & 0.447          & 0.166          & 0.459          & 0.217          & 0.474          & 0.125          & 0.545          & 0.412          & 0.605          & 0.366          & 0.662          \\
                                &\textit{ET}       & 0.562          & 0.721          & 0.305          & 0.720           & 0.150           & 0.473          & 0.093          & 0.450           & 0.188          & 0.312          & 0.167          & 0.331          \\ \midrule
\multirow{2}{*}{DeTime}         &\textit{ST}       & 0.145          & 0.254          & 0.123          & 0.254          & 0.038          & 0.028          & 0.031          & 0.038          & 0.050           & 0.100            & 0.050           & 0.100            \\
                                &\textit{ET}       & 0.511          & 0.602          & 0.176          & 0.274          & 0.054          & 0.142          & 0.029          & 0.038          & 0.059          & 0.088          & 0.051          & 0.075          \\ \midrule
PVTM                           &           & \textbf{0.722} & \textbf{0.744} & \textbf{0.755} & \textbf{0.765} & \textbf{0.366} & 0.569          & \textbf{0.595} & \textbf{0.766} & 0.583          & \textbf{0.787} & \textbf{0.825} & \textbf{0.817}
\\ \bottomrule 
\end{tabular}}
\caption{Text classification accuracy over 5-fold cross validation. The best results in each case are shown in \textbf{bold}.}
\label{tab:classification}
\end{table*}

%% file: tables/qualitative_result.tex
\begin{table*}[tp!]
\centering
\resizebox{0.8\linewidth}{!}{%
\begin{tabular}{@{}p{0.1\linewidth}l|l@{}}
\toprule
Models   & \begin{tabular}[c]{@{}l@{}}Topic Words\\ (on Short Text)\end{tabular}                                                                                         & \begin{tabular}[c]{@{}l@{}}Topic Words\\ (on LLAMA2-generated Long Text)\end{tabular}                                                                    
                                            \\ \midrule
LDA      & \begin{tabular}[c]{@{}l@{}}application,different,session,edit,\\ use,install,compile,long,design,setup\end{tabular}                     & \begin{tabular}[c]{@{}l@{}}app,library,use,build,cocoa,project, \\ application,dependency,framework,include\end{tabular}      \\ \midrule

ProdLDA     & \begin{tabular}[c]{@{}l@{}} apache, rewrite, redirect, url, log, \\ http, proxy, rule, php, domain \end{tabular}                     & \begin{tabular}[c]{@{}l@{}} database, oracle, datum, sql, query,   \\ table, data, linq, procedure,store
 \end{tabular}                                                            \\ \midrule

NMF     & \begin{tabular}[c]{@{}l@{}} oracle, value, way, select, row, \\ sql, table, database, column, procedure \end{tabular}                     & \begin{tabular}[c]{@{}l@{}} datum, data, large, allow, store \\ time, performance, need, ensure, work
 \end{tabular}                                                            \\ \midrule

BERTopic     & \begin{tabular}[c]{@{}l@{}} linq, sql, uery, join, use, \\ object, collection, expression, group, xml \end{tabular}                     & \begin{tabular}[c]{@{}l@{}} oracle, database, performance, row, statement, \\ table, sql, datum, query, use
 \end{tabular}                                                            \\ \midrule

NQTM     & \begin{tabular}[c]{@{}l@{}}image,come,null,application,pdf,\\hard,qstring,behave,repo,dynamically\end{tabular}                     & \begin{tabular}[c]{@{}l@{}} spring,application,development,framework,web\\
security,developer,platform,integrate,scalable \end{tabular}                                                            \\ \midrule

CTM      & \begin{tabular}[c]{@{}l@{}}cocoa,mac,app,os,application,\\osx,iphone,detect,development,audio\end{tabular}                             & \begin{tabular}[c]{@{}l@{}}spring,application,hibernate,configure,transaction,\\configuration,session,database,security boot\end{tabular}                                                                                          \\ \midrule

CLNTM    & \begin{tabular}[c]{@{}l@{}}mac,os,matlab,bash,command,\\qt,osx,context, url,rewrite\end{tabular}                                     & \begin{tabular}[c]{@{}l@{}}mac,app,os, apple,device,\\audio,video,cocoa,screen,quality\end{tabular}                \\ \midrule
TSCTM & \begin{tabular}[c]{@{}l@{}}eexample,axis,applescript,log,properly,\\ derive,hold,partition,line,spreadsheet\end{tabular} & \begin{tabular}[c]{@{}l@{}}studio,fxcop,visual,oslo,projects,\\awesome,editions,addon,eee,sharp\\ \end{tabular}                                                                                                                              \\ \midrule

vONT & \begin{tabular}[c]{@{}l@{}}oracle,cocoa,sql,datum,application,\\subversion,convert,different,select,xml\end{tabular} & \begin{tabular}[c]{@{}l@{}} branch,tuple,relational,orm,right,\\operator,standard,tree,trunk,left \end{tabular}

      \\ \midrule

DeTime & \begin{tabular}[c]{@{}l@{}}bash,sharepoint,page,class,table,\\string,load,line,variable,item\end{tabular} & \begin{tabular}[c]{@{}l@{}}shell,operator,icon,question,review,\\second,optimization,word,account,editor \end{tabular}

\\ \midrule

    PVTM      & \multicolumn{1}{c|}{-}                                                    & \begin{tabular}[c]{@{}l@{}}oracle, database, sql, table, store, \\ data, statement, procedure, query, index \end{tabular}  \\ \bottomrule 
\end{tabular}}
\caption{Topic words examples}
\label{tab:qualitative}
\vspace{-5mm}
\end{table*}

%% file: sections/conclusion.tex
\section{Conclusion}
In this paper, we address the issue of topic modeling for short texts. Our approach focuses on improving the input representation of short texts and enhancing the model's ability to capture latent topics despite the limited contextual information. The input to our method consists of individual short texts, such as a collection of tweets or headlines, and the output is a set of coherent topics that summarize the main themes present in the corpus. By tackling the data sparsity problem, we aim to develop a more effective topic modeling framework for short texts. A set of empirical evaluations demonstrate the effectiveness of the proposed framework over the state-of-the-art.

%% file: sections/acknowledgements.tex
\section*{Acknowledgements}
This material is based upon work supported by the National Science 
Foundation IIS 16-19302 and IIS 16-33755, Zhejiang University ZJU 
Research 083650, Futurewei Technologies HF2017060011 and 094013, 
IBM-Illinois Center for Cognitive Computing Systems Research (C3SR) and 
IBM-Illinois Discovery Accelerator Institute (IIDAI), grants from eBay 
and Microsoft Azure, UIUC OVCR CCIL Planning Grant 434S34, UIUC CSBS 
Small Grant 434C8U, and UIUC New Frontiers Initiative. Any opinions, 
findings, conclusions, or recommendations expressed in this publication 
are those of the author(s) and do not necessarily reflect the views of 
the funding agencies.

%% file: sections/limitation.tex
\section*{Limitations}
The proposed framework directly utilize LLMs for text generation conditioned on the given short texts. As we have specified before, this may result in noisy out-of-domain text generation, which hurts the document representativeness of the generated topics. This problem may worsen when the target domain is very specific. Although the proposed P-VTM tries to solve this problem, it does not work in extreme sparsity scenarios, as we observed in the TagMyNews dataset. Therefore, controlling the generation process such that it outputs more relevant text in the target domain is a possible future research direction in this line.

%% file: sections/appendix.tex
\appendix
\section{Appendix}
\input{tables/ablation}

\subsection{Implementation Details}
\label{sec:imp_details}

There are some parameters for both the proposed architecture and baselines we need to set. For text generation from LLMs. we use the maximum new tokens length as 500. We find that using beam-search decoding with a beam size of 5 generates more coherent text. The number of iterations for all the topic models is set to 100. For the smaller pretrained language model, we use SBERT\footnote{\href{https://huggingface.co/sentence-transformers
/paraphrase-distilroberta-base-v2}{https://huggingface.co/sentence-transformers/paraphrase-distilroberta-base-v2}} with a maximum sequence length of 512.
We use the huggingface library \footnote{\href{https://huggingface.co/docs/peft/en/package_reference/prefix_tuning}{https://huggingface.co/docs/peft/en/package\_reference/ prefix\_tuning}}  for prefix tuning with task type as FEATURE\_EXTRACTION and num\_virtual\_tokens as 20.
All parameters during calculating evaluation metrics are set to the same value across all the models. E.g., the number of top words for each topic for calculating $C_V$ and IRBO is set to 10. In text classification experiments, we use the default parameters for MNB from scikit-learn\footnote{\href{https://scikit-learn.org}{https://scikit-learn.org}}. For SVM, we use the hinge loss with the maximum iteration of 5. For logistic regression, the maximum iteration is set to 1000, and the tree depth for RF is set to 3 with the number of trees as 200.

\subsection{Input-Output Variants Analysis}
\label{sec:ablation}

In this section, we analyzed the effect of different input-output lengths on the performance of PVTM based on topic quality, as shown in \ref{sec:ablation}. The following variants of PVTM were considered: 
(1) \textbf{PVTM (S2S)}, which utilizes short texts for both input and output, (2) \textbf{PVTM (L2S)}, where long texts are used as input and short texts as output, (3) \textbf{PVTM (L2L)}, which operates on long texts for both input and output, and (4) \textbf{PVTM (S2L)}, our final model, which uses short texts as input and long texts as output.

The results demonstrate a clear improvement in topic quality ($C_V$, $IRBO$) across different versions, with the final variant showing the most significant gains. Replacing short texts entirely with generated long texts yields improvements over using only short texts. This enhancement is also evident when comparing the long-text-to-short-text variant. However, as previously noted, we selected short texts as input in our PVTM model for two main reasons: Semantic Consistency and Efficiency. Training with short texts helps mitigate the risk of introducing irrelevant information into the topic-modeling process. Although longer texts as the reconstruction objective may pose a risk of semantic drift, starting with short texts reduces this risk. Additionally, short texts enable faster processing, crucial for real-time topic detection. Using long texts as input would significantly increase inference time due to the LLM-processing overhead.




%% file: tables/ablation.tex
\begin{table*}[]
\centering
\resizebox{1.0\linewidth}{!}{%
\begin{tabular}{@{}l|cccccccccccc@{}}
\hline
\multicolumn{1}{l}{\multirow{3}{*}{Method}} & \multicolumn{4}{c}
{TagMyNews Titles}                                                                            & \multicolumn{4}{c}{Google News}                                                                                 & \multicolumn{4}{c}{StackOverflow}                                                                               \\ \cline{2-13} 
\multicolumn{1}{c}{}                        & \multicolumn{2}{c}{K=20}                               & \multicolumn{2}{c}{K=50}                               & \multicolumn{2}{c}{K=20}                               & \multicolumn{2}{c}{K=50}                               & \multicolumn{2}{c}{K=20}                               & \multicolumn{2}{c}{K=50}                               \\
\multicolumn{1}{c}{}                        & \multicolumn{1}{c}{$C_V$} & \multicolumn{1}{c}{$IRBO$} & \multicolumn{1}{c}{$C_V$} & \multicolumn{1}{c}{$IRBO$} & \multicolumn{1}{c}{$C_V$} & \multicolumn{1}{c}{$IRBO$} & \multicolumn{1}{c}{$C_V$} & \multicolumn{1}{c}{$IRBO$} & \multicolumn{1}{c}{$C_V$} & \multicolumn{1}{c}{$IRBO$} & \multicolumn{1}{c}{$C_V$} & \multicolumn{1}{c}{$IRBO$} \\ \midrule
PVTM (S2S)                   & 0.493 & 1.000 & 0.546 & 0.991 & 0.351 & 1.000 & 0.395 & 0.994 & 0.423 & 1.000 & 0.412 & 0.986                      \\ \midrule

PVTM (L2S)                & 0.515 & 1.000 & 0.497 & 0.989 & 0.357 & 1.000 & 0.401 & 0.994 & 0.519 & 1.000 & 0.432 & 0.998 \\
                                \midrule

PVTM (L2L)                 & 0.621 & 0.998 & 0.577 & 1.000 & 0.421 & 0.996 & 0.462 & 0.995 & 0.427 & 0.997 & 0.447 & 0.990                            \\ \midrule

\textbf{PVTM (S2L)}                & 0.632 & 1.000 & 0.585 & 1.000 & 0.445 & 1.000 & 0.445 & 1.000 & 0.452 & 1.000 & 0.462 & 1.000 \\
\bottomrule 
\end{tabular}
}
\caption{Performance of different variants of PVTM based on input and output lengths}
\label{tab:ablation}
\end{table*}